\documentclass[letterpaper, 10 pt, conference]{ieeeconf}  

\IEEEoverridecommandlockouts                              

\usepackage[T1]{fontenc}
\usepackage{bm}
\usepackage{graphicx}
\usepackage{caption}
\usepackage{subcaption}
\usepackage{multirow}
\usepackage{setspace}
\usepackage{amsmath}
\usepackage{booktabs} 
\usepackage{diagbox}
\usepackage{threeparttable}
\usepackage{hyperref}
\usepackage{makecell}

\title{\LARGE \bf
JPDS-NN: Reinforcement Learning-Based Dynamic Task Allocation for Agricultural Vehicle Routing Optimization
}

\author{Yixuan Fan$^{1,*}$, Haotian Xu$^{2,*}$, Mengqiao Liu$^{1}$, Qing Zhuo$^{1}$ and Tao Zhang$^{1,\dag}$, \textit {Fellow, IEEE}
\thanks{See our project page at \href{https://sites.google.com/view/jpds-nn-for-edvrp}{https://sites.google.com/view/jpds-nn-for-edvrp}.}.
\thanks{* Equal contribution. $\dag$ Corresponding author: taozhang@tsinghua.edu.cn}
\thanks{$^{1}$Department of Automation, Tsinghua University.}  \thanks{$^{2}$Beijing Institute of Astronautical Systems Engineering.}%
}

\begin{document}

\maketitle
\thispagestyle{empty}
\pagestyle{empty}

\begin{abstract}

The Entrance Dependent Vehicle Routing Problem (EDVRP) is a variant of the Vehicle Routing Problem (VRP) where the scale of cities influences routing outcomes, necessitating consideration of their entrances. This paper addresses EDVRP in agriculture, focusing on multi-parameter vehicle planning for irregularly shaped fields. 
To address the limitations of traditional methods, such as heuristic approaches, which often overlook field geometry and entrance constraints, we propose a Joint Probability Distribution Sampling Neural Network (JPDS-NN) to effectively solve the EDVRP.
The network uses an encoder-decoder architecture with graph transformers and attention mechanisms to model routing as a Markov Decision Process, and is trained via reinforcement learning for efficient and rapid end-to-end planning. 
Experimental results indicate that JPDS-NN reduces travel distances by 48.4–65.4\%, lowers fuel consumption by 14.0–17.6\%, and computes two orders of magnitude faster than baseline methods, while demonstrating 15–25\% superior performance in dynamic arrangement scenarios. Ablation studies validate the necessity of cross-attention and pre-training. The framework enables scalable, intelligent routing for large-scale farming under dynamic constraints.

\end{abstract}

\section{INTRODUCTION}

Autonomous farming systems are revolutionizing agriculture by boosting efficiency, tackling labor shortages, and promoting sustainability with precision technologies, ensuring food production amid climate and resource challenges \cite{LEE20102}. In the realm of unmanned farm operations, efficiently and economically allocating tasks to agricultural vehicles is of paramount importance. Numerous task-planning methods for agricultural vehicles have been proposed, many of which rely on dynamic programming algorithms and heuristic search approaches \cite{RN52, RN16, RN62}. However, these methods often fail to consider both the shape of the plots and the entrances of the working lines simultaneously.



\begin{figure}[h!]
    \centering
    \includegraphics[width=1\linewidth]{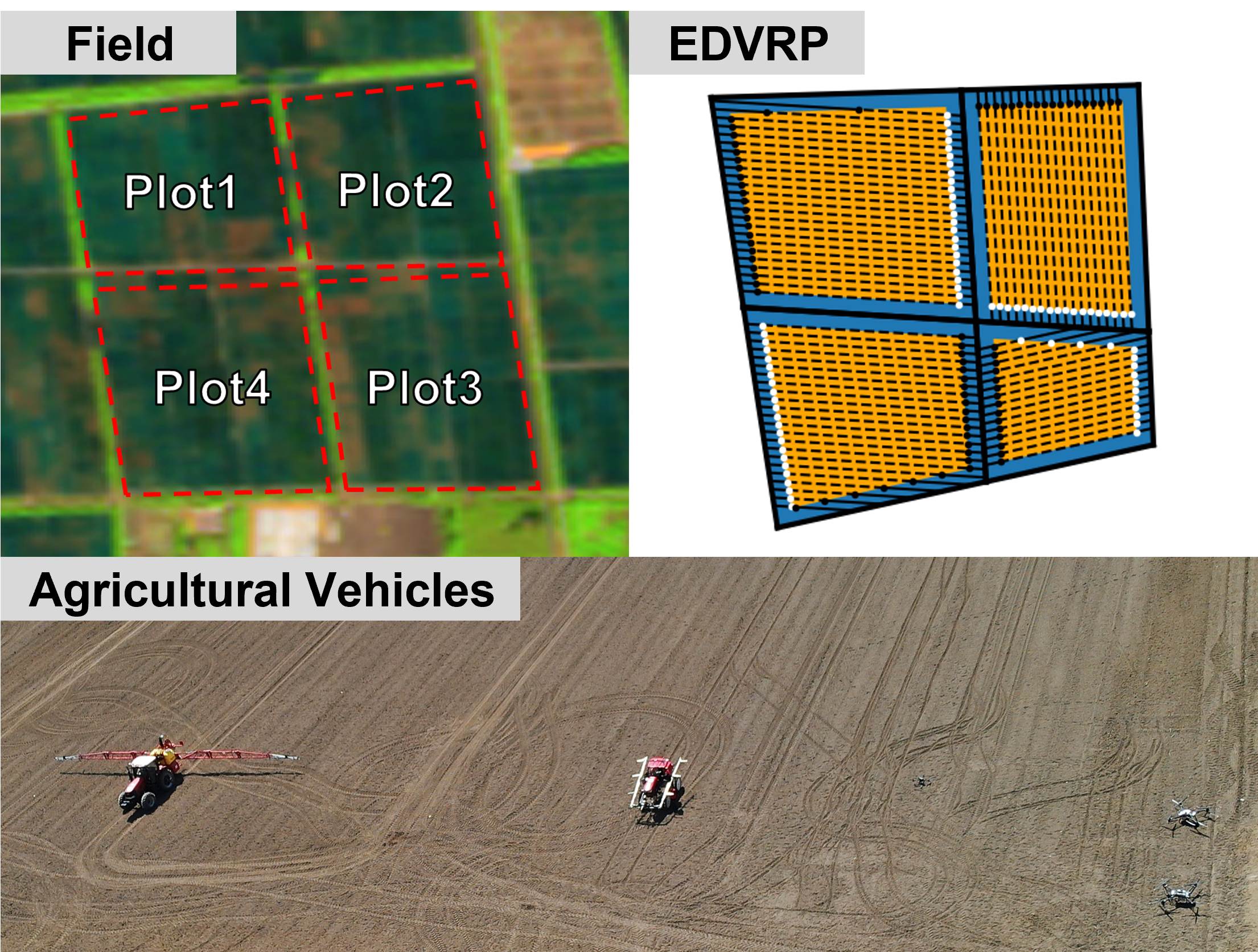}
    \caption{The EDVRP scenario in farms. 
    Top left: Real-world field with multiple plots. Bottom: Agricultural vehicles operate along working lines. Top right: Dashed lines represent working lines; white/black points represent entrances; solid lines represent the roads. Vehicles must start from the designated starting point, complete tasks without any redundancy, and then return to the ending point. The starting and ending points can either be the same location (such as a depot) or different locations (within working lines or roads).
    }
    \label{fig:EDVRP}
\vspace{-5mm}
\end{figure}

In our previous work \cite{OGA}, we defined the Entrance Dependent Vehicle Routing Problem (EDVRP) and presented a specific scenario of its application in farms, as illustrated in Fig. \ref{fig:EDVRP} . EDVRP is a specialized variant of the Vehicle Routing Problem (VRP) where the scale of cities is comparable to the distances between them, making city entrances a critical factor in route planning. Therefore, the optimization process in EDVRP must account for the entrances of each city. To address this, we introduced the Ordered Genetic Algorithm (OGA) for solving EDVRP in farm settings, which provided a solid baseline. However, like many heuristic algorithms, OGA is computationally intensive and often produces suboptimal results.

In this paper, we introduce a Joint Probability Distribution Sampling Neural Network (JPDS-NN) designed to solve EDVRP in farms more efficiently. JPDS-NN employs an encoder-decoder structure and models its inference process as a Markov Decision Process (MDP). The network includes an actor that samples actions from a joint probability distribution of working lines and their entrances. It is trained using Proximal Policy Optimization (PPO) \cite{ppo}. The main contributions of this paper are as follows:


\begin{itemize}
    \item We designed an encoder architecture based on graph transformers \cite{ijcai2021p214} and attention mechanisms \cite{RN119}, which facilitates rapid end-to-end task planning while effectively utilizing farm-specific information.
    \item We developed an actor network that samples actions from the joint probability distribution of working lines and their entrances, facilitating finer-grained task subdivision and enhancing the precision of agricultural vehicle operations.
    \item We implemented a simulator to visualize task allocation outcomes and designed dynamic arrangement tasks to validate the practicality and advancement of our method in real-world scenarios.
\end{itemize}

\section{RELATED WORKS}

In the early stages of research, heuristic search methods were commonly employed to address the VRP \cite{RN12, RN95, RN97, RN115, RN92}. In our previous work, we implemented an improved multi-mutation genetic algorithm \cite{OGA} to solve the EDVRP. However, among these methods, greedy algorithms often have a limited scope and can only achieve local optima. Heuristic search algorithms also tend to be computationally expensive, particularly when applied to large-scale problems.

O. Vinyals et al. \cite{RN2} introduced the idea that the Traveling Salesman Problem (TSP) can be modeled as a sequence-to-sequence problem, where the input is a sequence of cities and the output is a permutation of the input. Building on this concept, they proposed Pointer Networks (Ptr-Net) \cite{RN2}, where the network’s output serves as pointers to the input elements.

Given that VRP can be naturally represented as a graph, researchers have turned to Graph Neural Networks (GNNs) for extracting features from graph-structured data \cite{RN88, RN84, RN120}. L. Duan et al. \cite{RN120} developed a neural network that accounts for both nodes and edges, successfully solving VRP instances involving non-Euclidean distances.

In recent years, attention mechanisms \cite{RN116} have gained significant traction across various domains. W. Kool et al. \cite{RN86} applied Transformers to solve both TSP and VRP, while K. Lei et al. \cite{RN89} introduced the residual Edge-Graph Attention Network (residual E-GAT), which encodes node and edge features using a Graph Attention Network and outputs routing plans through a Transformer-based Ptr-Net. This method has outperformed others on both TSP and VRP tasks.

Reinforcement learning (RL) has been used to train networks for routing problems by exploring the state space and receiving rewards based on performance. I. Bello et al. \cite{RN5} applied RL algorithms to VRP, demonstrating better results than heuristic methods on large-scale TSPs, with significantly faster computational speed. Y. Bengio et al. \cite{RN87} argued that RL could potentially surpass expert decision-making and offers superior generalization capabilities. Consequently, RL has become one of the most popular training methods for VRP \cite{RN89, RN88, RN13}.

Currently, no known neural network-based methods have been proposed for solving EDVRP. To address this gap, we propose the Joint Probability Distribution Sampling Neural Network (JPDS-NN) for rapid end-to-end solutions in this paper.

\section{METHODOLOGY}
\subsection{Markov Decision Process for EDVRP in Farms}\label{MDP}

As described in our previous work\cite{OGA}, we represent the farm using a task graph $G = \langle \bm{N}, \bm{E} \rangle$, where $\bm{N}=\{\bm{n}_i|i=0,1,\dots,L\}$ is the set of nodes and $\bm{E}=\{\bm{d}_{ij}|i=0,1,\dots,L;j=0,1,\dots,L;i \neq j\}$ is the set of edges. Each node $\bm{n} \in \bm{N}$ represents a working line, and is defined as a vector $\bm{n} = [\cos \theta, \sin \theta, l]$, where $\theta$ is the angle between the working line and the east, and $l$ is the length of the line. 

To represent the starting and ending points of each vehicle, the first $M$ and the last $M$ nodes in the task graph are defined as $\bm{n}_{k}^{s} = \bm{n}_{k}^{e} = [0, 0, 0]$. Each starting node is unidirectionally linked to all working line nodes, which are also unidirectionally connected to ending nodes. Working line nodes are bidirectionally interconnected, with no direct connections between start and end points. This setup shows that each vehicle's task starts at its origin, passes through working lines, and ends at its destination.

The edge $\bm{d}_{ij} \in \bm{E}$, between nodes $\bm{n}_i$ and $\bm{n}_j$, is represented as $\bm{d}_{ij} = \left[ d^*_{ij00}, d^*_{ij01}, d^*_{ij10}, d^*_{ij11} \right]$, where $d^*_{ij m_i m_j}$ is the shortest available distance between entrance $m_i$ of working line $i$ and entrance $m_j$ of working line $j$. 

Each vehicle $k$ is characterized by its working speed $v_k^w$, idle speed $v_k^f$, fuel consumption during operation $c_k^w$, and fuel consumption during idling $c_k^f$. This is represented as $\bm{m}_k = \left[v_k^w, v_k^f, c_k^w, c_k^f\right]$.

\begin{figure}[t]
  \centering
  \includegraphics[width=1\linewidth]{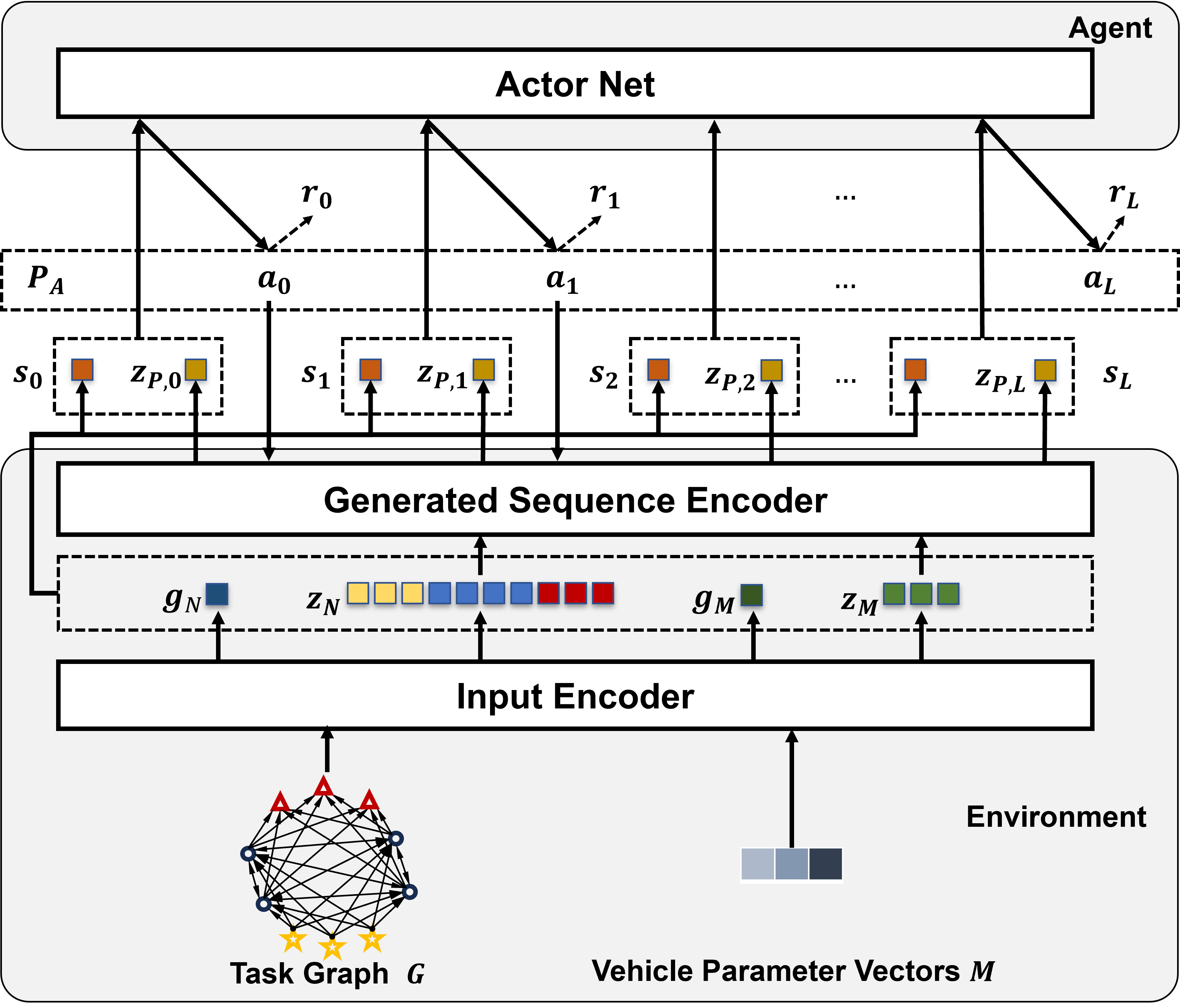}  
  \caption{The input encoder handles the task graph and vehicle features. The decoder, comprising a sequence encoder, actor network, and critic network, produces a sequence. In the MDP, the environment includes inputs, input encoder, and sequence encoder, with the actor network as the agent. The input encoder extracts high-dimensional input features, and the sequence encoder processes action features. At each step, the actor network chooses an action \( a = P_{A,t} \) based on the state, determining the next node and its entrance.}
  \label{fig:MDP}
\end{figure}

The inputs to our network include the task graph $\bm{G}$ and the vehicle parameter vectors $\bm{M} = \{\bm{m}_k | k = 1, \dots, M\}$, while the output is a sequence of node-entrance pairs $\bm{P}_A$. The nodes in $\bm{P}_A$ form a permutation of all nodes in the task graph, which includes working lines, starting points and ending points. The starting and ending points serve to divide the sequence $\bm{P}_A$ into $M$ segments, with each segment representing the operation sequence of a specific vehicle. We describe this process using a MDP, as shown in Fig. \ref{fig:MDP}. 

\begin{figure*}[t]
  \centering
  \includegraphics[width=0.99\linewidth]{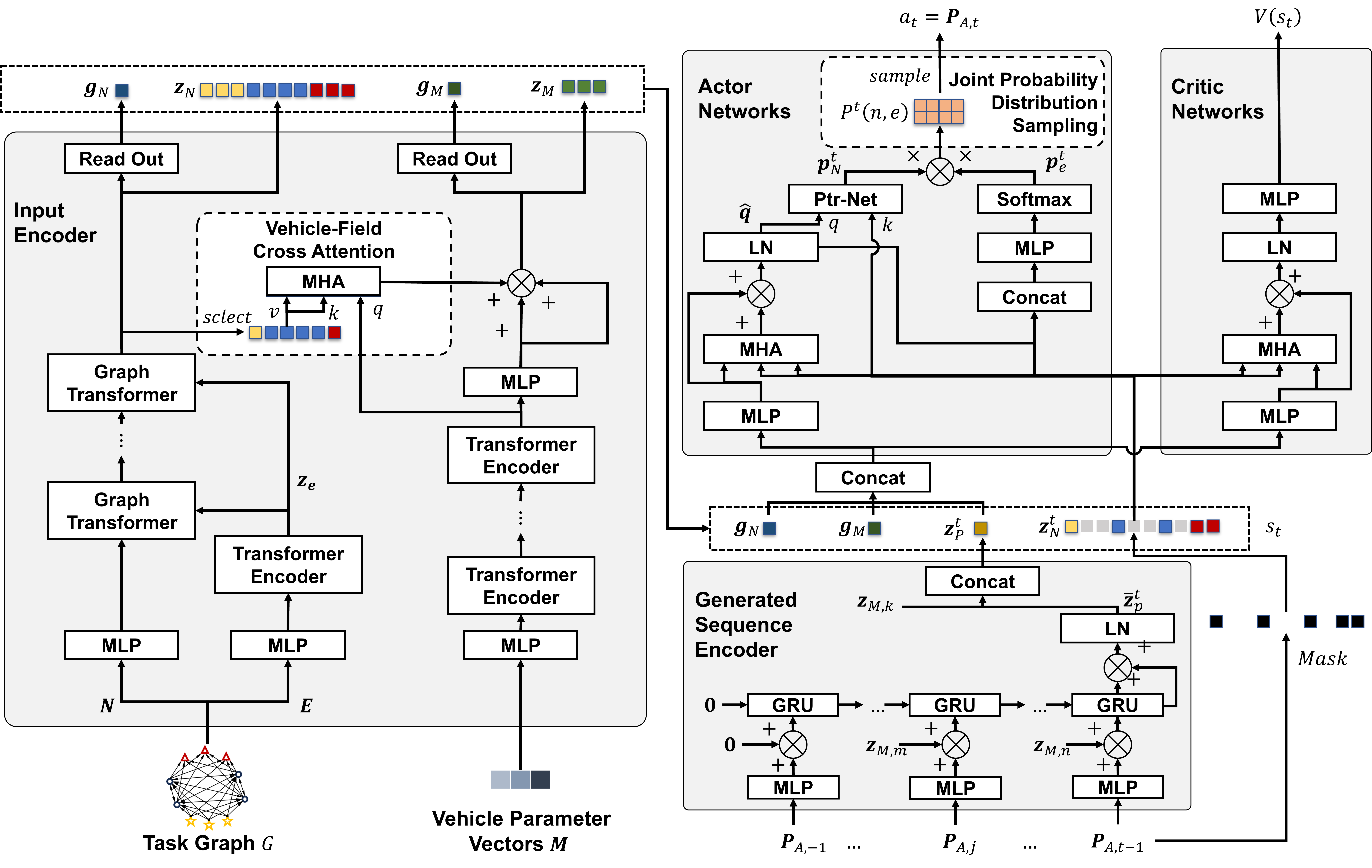}
  \caption{Detailed structure of our networks. In the figure, MLP refers to Multi-Layer Perceptron; Attn refers to Attention; MHA refers to Multi-Head Attention\cite{RN119}; LN refers to Layer Normalization; Concat refers to vector concatenation; GRU refers to Gated Recurrent Unit\cite{gru}; and Read Out concatenates the bitwise maximum and average values of a sequence, using MLP to produce a global feature.}
  \label{fig: encoder}
\end{figure*}

In Ref. \cite{OGA}, we defined three key objectives: total distance, maximum time, and total fuel consumption of the vehicles. In this work, we define the reward for each MDP step based on the change in these objectives. We note the actions at step $t-1$ and $t$ are $(i,e_i)$ and $(j, e_j)$, respectively, where in each action the former element is the selected node and the latter one is its selected entrance. 

For distance, the reward at step $t$ is:
\begin{equation}
    r^s_t = d^*_{ij\bar{e}_i e_j}.
\end{equation}

Let the operation sequence of vehicle $m$ after step $t$ be $p_m$, where $m = 1, \dots, M$, and let the sequence length be $n_m$. The $h$-th element represents node $\bm{n}_{\hat{h}}$ and entrance $e_{\hat{h}}$. The time consumption for vehicle $m$ is:
\begin{equation}
    t_m = \frac{1}{v^f_m} \sum_{h=1}^{n_m-1} d^*_{\hat{h}(\hat{h}+1)\bar{e}_{\hat{h}} e_{\hat{h}+1}} + \frac{1}{v^w_m} \sum_{h \in p_m} l_{\hat{h}}. 
\end{equation}

After step $t$, a new node is added to vehicle $k$'s operation sequence $p_k$. The new sequence is denoted as $p'_k$, and the updated time consumption for vehicle $k$ is:
\begin{equation}
    t'_k = \frac{1}{v^f_k} \sum_{h=1}^{n_k} d^*_{\hat{h}(\hat{h}+1)\bar{e}_{\hat{h}} e_{\hat{h}+1}} + \frac{1}{v^w_k} \sum_{h \in p'_k} l_{\hat{h}}. 
\end{equation}

Thus, the time increment due to action $a_t$ is:
\begin{equation}
    r^t_t = \max\left(t'_k - \max_{m} t_m, 0 \right).
\end{equation}

For fuel consumption, the increment from action $a_t$ is:
\begin{equation}
    r^c_t = \frac{1}{c^f_k} d^*_{ij\bar{e}_i e_j} + \frac{1}{c^w_k} l_j.
\end{equation}

Considering the relationships among the three objectives, we design a method to combine rewards. When training networks optimized for time or fuel consumption, we add $r_t^t$ to either $r_t^s$ or $r_t^c$, depending on the specific goal. We call the additional reward the distance bonus. 

\subsection{Joint Probability Distribution Sampling Neural Network}

Our neural network consists of an input encoder, a generated sequence encoder, actor networks, and critic networks, as illustrated in Fig. \ref{fig: encoder}.

\subsubsection{Input Encoder}

The input encoder processes the inputs to extract high-dimensional features, generating feature sequences \(\bm{z}_N\) for nodes in the task graph and \(\bm{z}_M\) for vehicles, along with their corresponding global feature vectors \(\bm{g}_N\) and \(\bm{g}_M\). An 8-layer Graph Transformer \cite{ijcai2021p214} is employed to extract node features from the task graph. Given the presence of both unidirectional and bidirectional connections in the task graph, extracting edge features is crucial. To address this, a Transformer Encoder block \cite{RN119} is utilized to capture edge cross-correlation features using a self-attention mechanism, as the relationships between edges significantly impact the model's performance. For vehicle feature extraction, 2-layer Transformer Encoder Blocks and Cross-Attention mechanisms are applied, considering the importance of cross-correlation features between vehicles and between vehicles and the task graph. In the Vehicle-Field Cross Attention block, vehicle features act as the query, while node features serve as the key and value. Dynamic mask generation ensures that each vehicle feature focuses exclusively on its designated starting and ending nodes. 

We designed a pre-training task for node clustering for Graph Transformer layers. Nodes belonging to the same plot in the task graph are labeled as one category, while the starting and ending points are labeled as two other categories. The goal of the clustering task is to classify nodes within the same plot into one category and distinguish them from nodes in other plots. We employ Binary Cross-Entropy Loss for pre-training:
\begin{equation}
    \text{Loss} = -\frac{1}{N} \sum_{i=1}^{N} \left[ y_i \cdot \log(p_i) + (1 - y_i) \cdot \log(1 - p_i) \right],
\end{equation}

where $N$ is the number of samples, $y_i$ is the true label of the $i$-th sample, $p_i$ is the predicted probability of the positive class.

\subsubsection{Generated Sequence Encoder}
A node-entrance pair in $\bm{P}_A$ is encoded by combining the node's feature with a one-hot vector of the entrance. For example, the pair $\bm{P}_{A,j} = (\bm{n}_i, 1)$ is encoded as $[\bm{z}_{N,i}, 0, 1]$. $\bm{P}_{A,-1}$ represents the depot. The generated sequence encoder utilizes a single-layer GRU to process the generated sequence, producing a feature vector $\bm{z}_P^t$. We enhance the network's ability to summarize past decisions by adding the corresponding vehicle’s feature to each action. The current vehicle’s feature $\bm{z}_{M,k}$ is concatenated with the output, as it is critical for making the current decision. 

\subsubsection{Actor Networks}
The actor networks first encode the state. The generated sequence is used to create a mask for the nodes, preventing nodes that have been selected before step $t$ from being chosen again. The depot is masked after it has been selected $M-1$ times. The masked node sequence, along with $\bm{g}_N$, $\bm{g}_M$, and $\bm{z}_P^t$, are encoded into a state feature $\hat{\bm{q}}$ using a multi-head attention mechanism, similar to the approach in Ref. \cite{RN89}. The concatenation of $\bm{g}_N$, $\bm{g}_M$, and $\bm{z}_P^t$ serves as the query, while the masked node features are used as the key and value. This state feature $\hat{\bm{q}}$ and the masked node sequence are passed as the query and key to a Ptr-Net\cite{RN2}, which outputs the probabilities for selecting each unchosen node. 

Since the selection of an entrance for each node is also required, we develop an entrance probability network. Let the feature of a node in the masked node sequence be $\bm{z}_{N,w}^t$. The probabilities for choosing each entrance are calculated as:
\begin{equation}    
\bm{p}_{e,w}^t = \text{softmax}\left(\text{MLP}\left(\hat{\bm{q}}; \bm{z}_{N,w}^t\right)\right),
    \label{eq:entry_prob}
\end{equation}
where $\bm{p}_{e,w}^t$ is a 2-dimensional vector representing the probabilities of selecting entrances 0 or 1. 

By multiplying the node probability with the corresponding entrance probability, we obtain the joint probability distribution $P^t(n,e)$ for selecting node $i$ and its entrance $e_i$. If the $i$-th node in the task graph corresponds to the $w$-th item in the unselected node sequence, the probability of the actor network choosing the action $a_t = P_{A,t} = (i, e_i)$ is given by:
\begin{equation}
    p(a_t | s_t) = p_{N,w}^t \times p_{e,w,e_i}^t.
\end{equation}
The actor network samples actions from this joint probability distribution.

\subsubsection{Critic Networks}
The critic networks are similar in structure to the actor networks. They use MLPs to process the state feature and produce a value estimate $V(s_t)$. While the critic and actor networks share some architectural similarities, they do not share parameters. 

\section{EXPERIMENTS}




In this section, we assess the performance of JPDS-NN in optimizing task allocation for the EDVRP. The experiments are designed to achieve four key objectives: 

\begin{itemize}
    \item Evaluating the performance of the designed networks during the training process across three optimization objectives;
    \item Comparing JPDS-NN, under the three optimization objectives, with baseline methods on the test set;
    \item Conducting ablation studies to verify the effectiveness of pre-training and network feature design;
    \item Designing two dynamic arrangement tasks to further evaluate the method's performance in real-world scenarios.
\end{itemize}

\subsection{Experimental Setup}

All networks were trained using Proximal Policy Optimization (PPO) algorithm \cite{ppo}, with a batch size of 64, learning rate of 0.0001, dropout rate of 0.1, and discount factor \(\gamma\) of 0.99. In the network implementation, the number of attention heads in all multi-head attention modules is set to 4. The suffixes -s, -t, and -c correspond to the three optimization objectives: total distance \( s_P = \sum_{k=1}^{M} s_k \), maximal time \( t_P = \max_k t_k \), and total fuel consumption \( c_P = \sum_{k=1}^{M} c_k^o \), respectively. 

The training, validation, and test sets we used contain 10,800, 1,080, and 7,500 scenarios, respectively, with each scenario comprising {2, 3, 4, 5, 6} plots and {2, 3, 4, 5, 6} vehicles. The vehicle's working speed \( v^w \) ranges randomly from 1 m/s to 3.3 m/s. The empty travel speed \( v^f \) ranges from \( \max(v^w, 2) \) m/s to 6.94 m/s, ensuring a minimum empty travel speed of 2 m/s and \( v^f \geq v^w \), with a maximum of 25 km/h. The empty travel fuel consumption \( c^f \) ranges randomly from 0.005 L/s to 0.008 L/s, while the working fuel consumption \( c^w \) ranges from \( \max(c^f, 0.007) \) L/s to 0.01 L/s, ensuring \( c^w \geq c^f \). 


\begin{figure*}[h!]
    \centering
    \begin{subfigure}{0.325\textwidth}
        \centering
        \includegraphics[width=\linewidth]{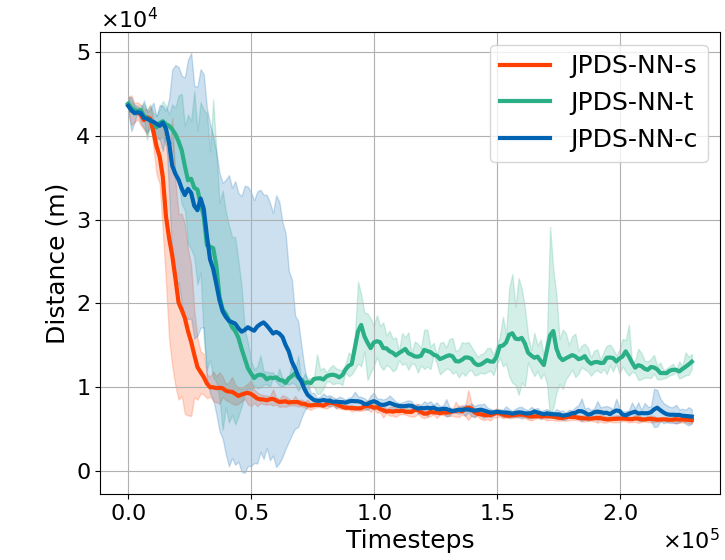}
    \end{subfigure}
    \begin{subfigure}{0.325\textwidth}
        \centering
        \includegraphics[width=\linewidth]{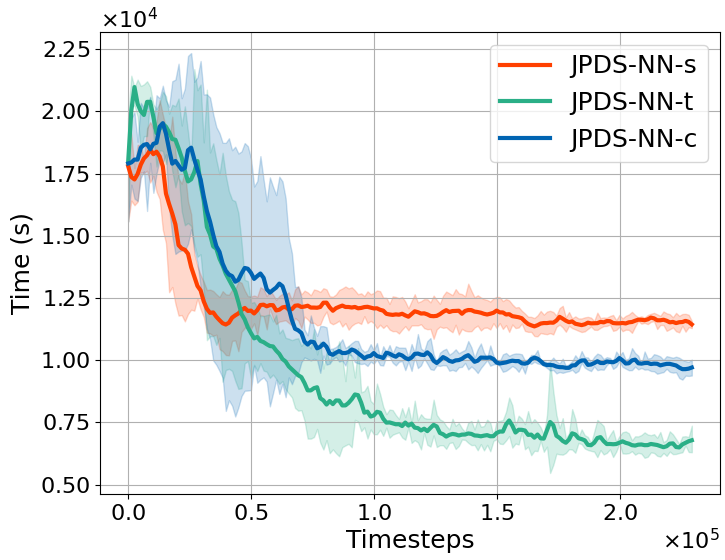}
    \end{subfigure}
    \begin{subfigure}{0.325\textwidth}
        \centering
        \includegraphics[width=\linewidth]{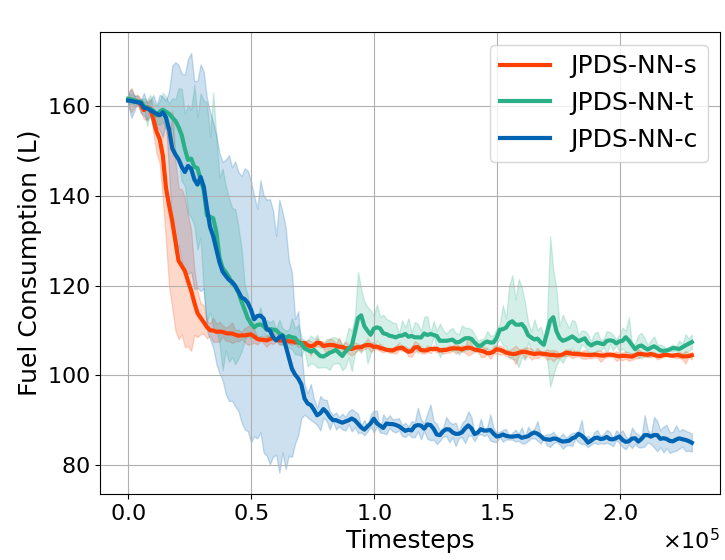}
    \end{subfigure}
    \caption{Training curves of JPDS-NNs under four random seeds. In the early stages, the optimization of distance, time, and fuel consumption aligns, but with training processes, these objectives may diverge or conflict. In the later stages, as the algorithm provides better allocation, the optimization directions converge again.}
    \label{fig:training_curve}
\end{figure*}

\subsection{Training Results}

The training curves in Fig. \ref{fig:training_curve} demonstrate the convergence and effectiveness of JPDS-NN. It can be seen that optimizing one objective simultaneously improves the other two, indicating a coupling between the three objectives. When one objective is set as the optimization target, the network provides a solution that significantly outperforms the scenario where it is not the target, showing that while the three objectives are correlated, they do not reach optimality simultaneously.

Among the three optimization objectives, distance is fundamental. Optimizing distance relates only to the task graph, not vehicle characteristics. While optimizing one vehicle's distance can positively impact its total distance and running time, it doesn't always improve overall fleet time or fuel consumption. After 40,000 training scenarios of JPDS-NN-s, time increases slightly, while fuel consumption and distance both decrease, suggesting that early distance optimization helps both time and fuel efficiency. However, further optimizing distance alone can negatively impact time and only slightly improve fuel efficiency.

In Section \ref{MDP}, we introduced a method to combine rewards, demonstrating that optimizing for time and fuel consumption can also enhance operational distance. To achieve this, a distance term \( r^s \) was added to the reward function during the first 7 training epochs (75,264 timesteps) for the time-optimized objective and throughout the entire training process for the fuel consumption-optimized objective. As depicted in Fig. \ref{fig:training_curve}, extended training focused on these objectives leads to a reduction in distance as well.

\subsection{Baseline Comparison}
In this section, we evaluated the performance of JPDS-NN against two baselines: RA and OGA \cite{OGA}. Here, RA refers to Random Arrangement, which serves as a baseline representation of the task scenario's complexity. OGA, on the other hand, is an enhanced genetic algorithm we previously proposed, specifically tailored to solve the EDVRP problem.

\begin{table}[h]
\caption{The results of JPDS-NN and baseline methods on the test set. Best results in bold.} \label{tab:baseline_oga}
\small
\centering
\begin{tabular}{w{l}{1.7cm}w{c}{1.2cm}w{c}{1.2cm}w{c}{1.2cm}w{c}{1.2cm}}
\toprule
Approach &  Distance (m) & Time (s) & Fuel (L) & Runtime (s) \\ \midrule
RA     & $37684.36$  & $10429.99 $& $140.93$ & - \\ \hline \noalign{\vskip 1mm}
OGA-s       & $15934.20$  & $\textbf{7144.13}$ & $107.44$ & $14.19$ \\
JPDS-NN-s   & $\textbf{5701.28}$  & $8852.70$ & $\textbf{91.90}$ & $\textbf{0.10}$\\ \hline \noalign{\vskip 1mm}
OGA-t       & $21185.98$  & $4178.40$ & $108.62$ & $14.19 $\\
JPDS-NN-t   & $\textbf{10941.95}$  & $\textbf{3928.48}$ & $\textbf{93.36}$ & $\textbf{0.10}$ \\ \hline \noalign{\vskip 1mm}
OGA-c       & $18350.94$  & $8558.72$ & $85.32$ & $14.04$ \\  
JPDS-NN-c   & \textbf{6353.10}  & \textbf{8101.46} & \textbf{70.28} & \textbf{0.07} \\
\bottomrule
\end{tabular}
\end{table}

As illustrated in Table \ref{tab:baseline_oga}, JDPS-NN significantly outperforms the genetic algorithm in travel distance optimization. When optimized for travel distance, time, and fuel consumption, JPDS-NN achieves 64.2\%, 48.4\%, and 65.4\% shorter average travel distances, respectively, compared to OGA on test set. Optimizing for fuel consumption often requires fewer machines, leading to a strong correlation with travel distance. Consequently, JPDS-NN surpasses the genetic algorithm in fuel consumption by 14.5\%, 14.0\%, and 17.6\% for each respective metric. 

Comparing fuel consumption results in the table, JPDS-NN-c achieves 23.5\% lower fuel consumption than JPDS-NN-s, despite a 14.9\% longer travel distance. This shows the network can select vehicles with lower fuel consumption, reducing overall fleet fuel usage.

JPDS-NN-t improves travel time optimization by only 5.9\% over the genetic algorithm but significantly reduces travel distance and fuel consumption. This indicates the neural network excels in optimizing task sequences for individual machines but is less effective in allocating tasks across multiple vehicles.


In terms of computational efficiency, JPDS-NN significantly outperforms OGA, making it an ideal choice for large-scale farming applications and dynamic machinery dispatch during operations, particularly in response to shifting demands or equipment breakdowns.

\subsection{Ablation Study\label{ablation}}

\subsubsection{Ablation of pre-training}

\begin{figure}[h!]
    \centering
    \begin{subfigure}{0.49\linewidth}
        \centering
        \includegraphics[width=\linewidth]{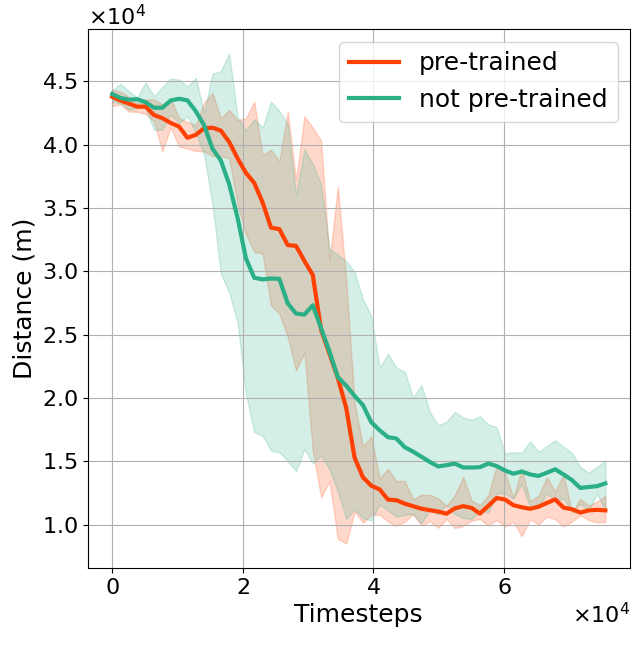}
    \end{subfigure}
    \begin{subfigure}{0.49\linewidth}
        \centering
        \includegraphics[width=\linewidth]{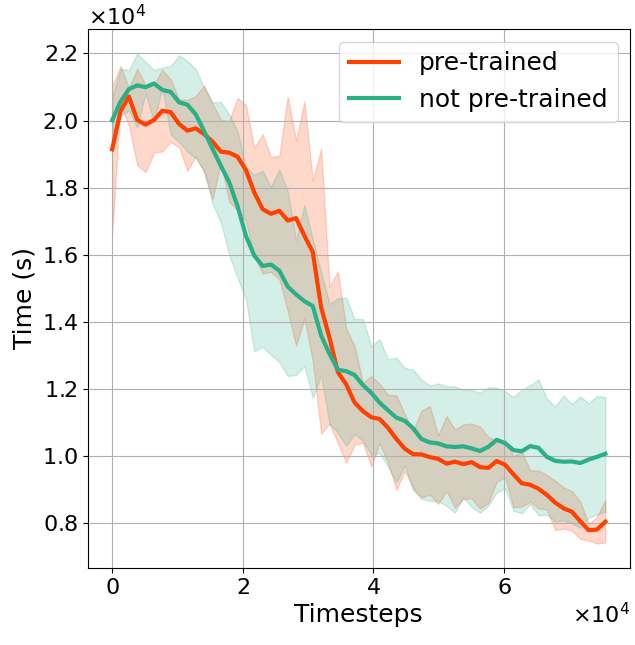}
    \end{subfigure}
    \caption{The network without pre-training converges as quickly or even faster in the early training stages. However, as training progresses, the pre-trained network shows less fluctuations and a better performance on the validation set.}
    \label{fig:pretrain}
\vspace{-1mm}
\end{figure}

This section verifies the significance of the pre-trained graph encoder for network training. Noted that the graph encoder solely impacts node encoding, and since distance optimization depends only on node encoding, we evaluates the effectiveness of using pre-trained graph encoders in the training process of JPDS-NN-s. Results are presented in Fig. \ref{fig:pretrain}. 

The network without pre-training shows inferior training stability and performance compared to the pre-trained network. A randomly initialized network can initially achieve some optimization capability for task planning, quickly converging to relatively optimal distances as the decoder trains. However, The pre-trained encoder extracts node features that enable shorter distances, while the network without pre-training extracts inferior features, leading to better optimality for the pre-trained network in later training stages. On the other hand, the pre-trained encoder has already converged to a relatively stable state, which helps the network achieve greater stability during subsequent training.

\subsubsection{Ablation of cross-attention}


In Section \ref{MDP}, we defined independent starting and ending points \(\{\bm{n}_{k}^{s}, \bm{n}_{k}^{e} | k = 1, \dots, M\}\) for each vehicle. However, in most real-world scenarios, each field has only one depot, from which all vehicles depart and return after completing their tasks. In such cases, the start and end points in the task graph can be simplified to a single depot node \(\bm{n}_{d}\). Consequently, the Vehicle-Field Cross Attention block in the encoder can also be removed. To evaluate this, we trained a simplified network optimized for time (denoted as JPDS-NN-sd) and tested it on the single-depot scenario (consisting of 2,500 scenarios) from the test set. 

\begin{table}[h]
\caption{The results on single-depot scenarios.} \label{tab:ablation_ia}
\small
\centering
\begin{tabular}{w{l}{1.7cm}w{c}{1.2cm}w{c}{1.2cm}w{c}{1.2cm}w{c}{1.2cm}}
\toprule
Approach &  Distance (m) & Time (s) & Fuel (L) \\ \midrule
JPDS-NN-sd       & $11147.21$  & $4226.10$ & $93.98$\\  
JPDS-NN   & \textbf{11098.84}  & \textbf{4045.93} & \textbf{93.91}\\
\bottomrule
\end{tabular}
\label{tab:compare_ia}
\end{table}

The results are shown in Table \ref{tab:ablation_ia}. It is clear that the cross-attention mechanism not only enables the network to encode the start and end points of each vehicle in the task graph but also significantly improves the network's performance.

\subsection{Dynamic Arrangement}

In real-world scenarios, tasks may change during execution, requiring adaptive re-arrangement. Our network supports single-round planning based on the task graph and vehicle information and dynamically optimizes task allocation when conditions change. Therefore, we designed two dynamic arrangement tasks based on real-world agricultural production needs, as described in the following.

\noindent \textbf{Field Increase.} When plots within a field cannot all be worked on simultaneously (e.g., certain plots require priority or are temporarily unavailable), it becomes necessary to plan for a subset of plots first and then incrementally add and re-arrange for additional plots. 

\noindent \textbf{Vehicle Decrease.} During operations, agricultural vehicle may be required to return to the depot (e.g., for refueling or unloading crops). In such cases, the remaining tasks need to be reassigned to other vehicles.


We performed simulation experiments on the test set for both tasks using JPDS-NN and OGA. The experimental design was structured as follows: after the task has been underway for a period of time, additional plots were introduced or vehicles were removed, and a re-arrangement was subsequently carried out. Metrics were recorded upon the completion of the second task. The experimental results are detailed in Table \ref{tab:da}. 

\begin{table}[h]
\caption{The results of dynamic arrangement experiment} 
\label{tab:da}
\small
\centering
\begin{tabular}{w{c}{1.4cm}w{l}{1.5cm}w{c}{1.2cm}w{c}{1.2cm}w{c}{1cm}}
\toprule
Setting &  Approach &  Distance (m) & Time (s) & Fuel (L) \\ \midrule
&OGA-s       & $21274.02$  & $7503.50$ & $69.10$ \\
&JPDS-NN-s   & $\textbf{17576.12}$  & $\textbf{5847.22}$ & $\textbf{62.74}$ \\ \cline{2-5} \noalign{\vskip 1mm}
Field&OGA-t       & $27015.14$  & $2886.22$ & $70.16$ \\
Increase&JPDS-NN-t   & $\textbf{21233.26}$  & $\textbf{2773.26}$ & $\textbf{61.85}$ \\ \cline{2-5} \noalign{\vskip 1mm}
&OGA-c       & $24206.17$  & $6274.82$ & $56.16$ \\  
&JPDS-NN-c   & $\textbf{18195.32}$  & $\textbf{5834.96}$ & $\textbf{50.89}$ \\ \noalign{\vskip 1mm} \hline \hline \noalign{\vskip 1mm}
&OGA-s       & $22569.49$  & $8015.20$ & $71.53$ \\
&JPDS-NN-s   & $\textbf{19011.60}$  & $\textbf{6614.87}$ & $\textbf{66.36}$ \\ \cline{2-5} \noalign{\vskip 1mm}
Vehicle&OGA-t       & $27311.02$  & $4672.96$ & $76.22$ \\
Decrease&JPDS-NN-t   & $\textbf{21966.99}$  & $\textbf{4527.76}$ & $\textbf{68.40}$ \\ \cline{2-5} \noalign{\vskip 1mm}
&OGA-c       & $25799.18$  & $7954.80$ & $65.37$ \\  
&JPDS-NN-c   & $\textbf{20750.69}$  & $\textbf{7038.42}$ & $\textbf{60.44}$ \\
\bottomrule
\end{tabular}
\end{table}

The experimental results are consistent with those in Table \ref{tab:baseline_oga}, validating the applicability and superiority of our method in real-world scenarios. In both tasks, JPDS-NN outperforms OGA across all metrics for the three optimization objectives. This can be attributed to the fact that during re-arrangement, vehicles may be located within working lines, on transfer paths, or in the depot. At this point, JPDS-NN can extract the relative positional relationships of the vehicles within the overall field, thereby facilitating more optimized task allocation decisions.



\begin{figure*}[t]
    \centering
    \begin{subfigure}{0.325\textwidth}
        \centering
        \includegraphics[width=\linewidth]{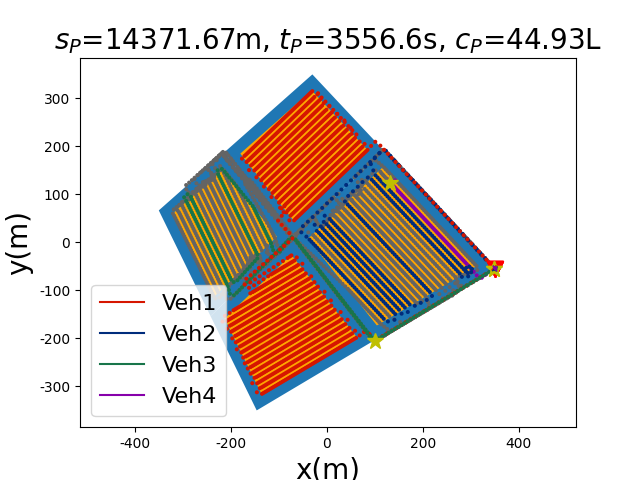}
        \caption{\small JPDS-NN-s}
    \end{subfigure}
    \hfill 
    \begin{subfigure}{0.325\textwidth}
        \centering
        \includegraphics[width=\linewidth]{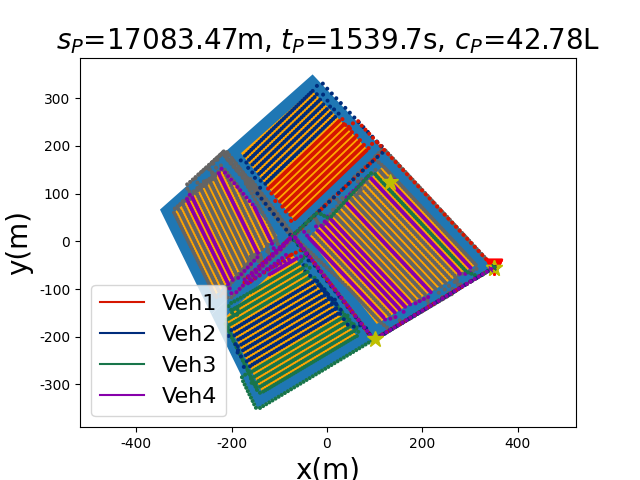}
        \caption{\small JPDS-NN-t}
    \end{subfigure}
    \hfill 
    \begin{subfigure}{0.325\textwidth}
        \centering
        \includegraphics[width=\linewidth]{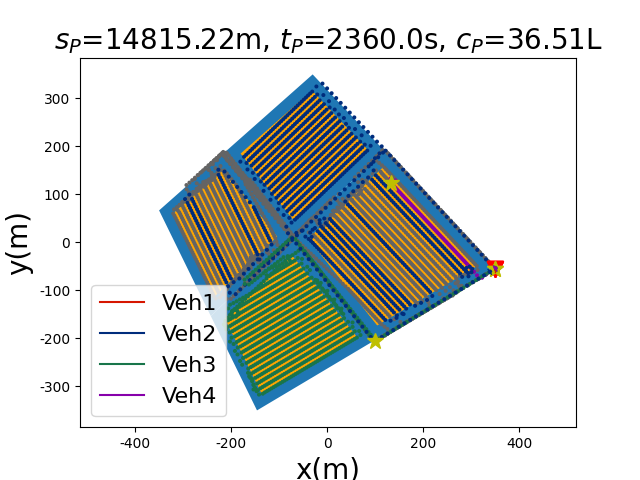}
        \caption{\small JPDS-NN-c}
    \end{subfigure}

    \begin{subfigure}{0.325\textwidth}
        \centering
        \includegraphics[width=\linewidth]{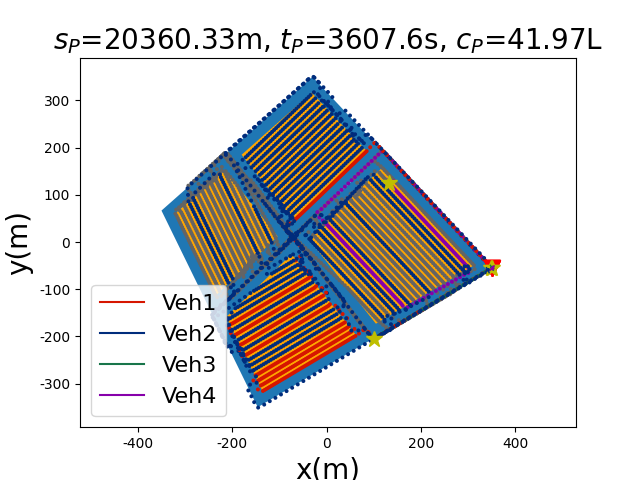}
        \caption{\small OGA-s}
    \end{subfigure}
    \hfill 
    \begin{subfigure}{0.325\textwidth}
        \centering
        \includegraphics[width=\linewidth]{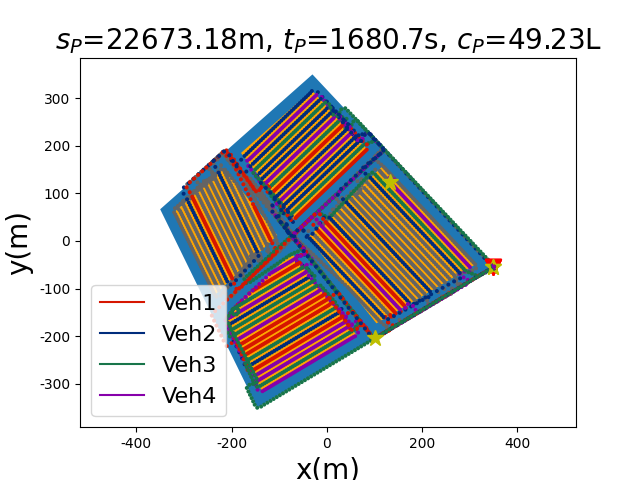}
        \caption{\small OGA-t}
    \end{subfigure}
    \hfill 
    \begin{subfigure}{0.325\textwidth}
        \centering
        \includegraphics[width=\linewidth]{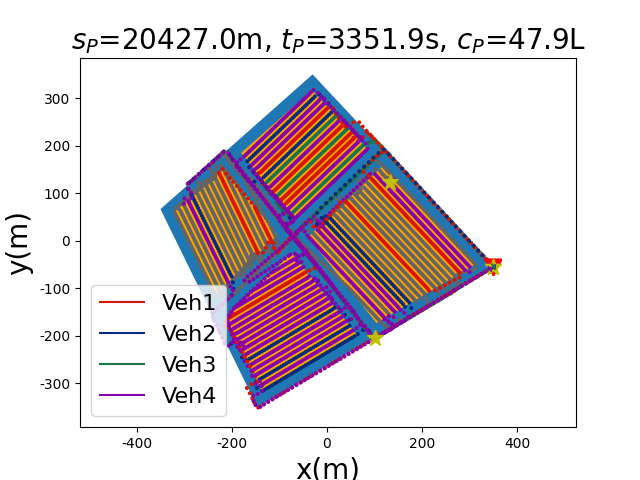}
        \caption{\small OGA-c}
    \end{subfigure}\label{fig:case_study_1}
    \caption{Simulation results of filed increase task. In the first phase, all four vehicles depart from the depot and operate on Plot 0 (top-left) and Plot 3 (bottom-right). At 50\% of the operation time, Veh1 and Veh3 have returned to the depot, Veh2 is on a transfer path, and Veh4 is within a working line. In the second phase, the remaining plots were added, and the vehicles commenced their operations from their respective starting points, eventually returning to the depot.}
    \label{fig:case_study_field}
\vspace{-1mm}
\end{figure*}

\begin{figure*}[h]
    \centering
    \begin{subfigure}{0.325\textwidth}
        \centering
        \includegraphics[width=\linewidth]{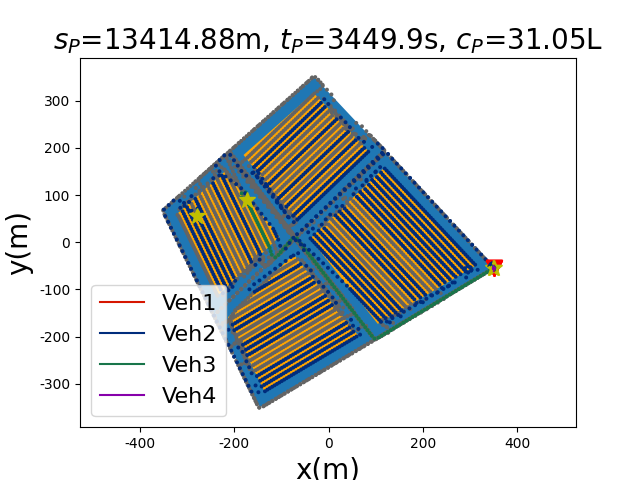}
        \caption{\small JPDS-NN-s}
    \end{subfigure}
    \hfill 
    \begin{subfigure}{0.325\textwidth}
        \centering
        \includegraphics[width=\linewidth]{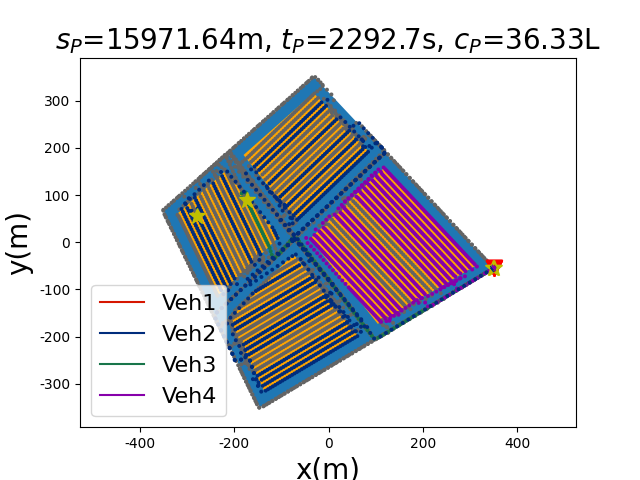}
        \caption{\small JPDS-NN-t}
    \end{subfigure}
    \hfill 
    \begin{subfigure}{0.325\textwidth}
        \centering
        \includegraphics[width=\linewidth]{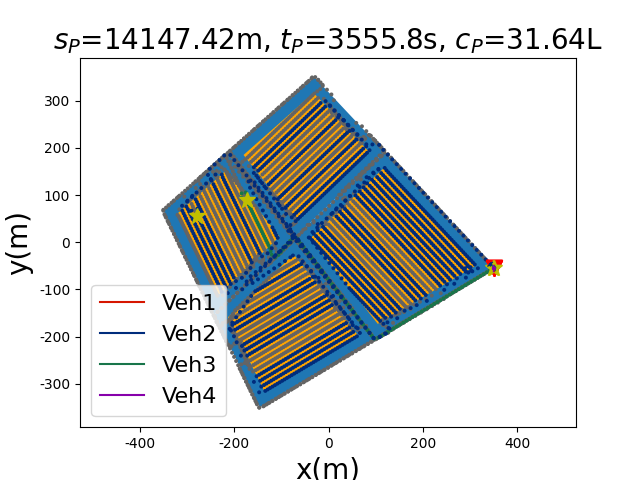}
        \caption{\small JPDS-NN-c}
    \end{subfigure}

    \begin{subfigure}{0.325\textwidth}
        \centering
        \includegraphics[width=\linewidth]{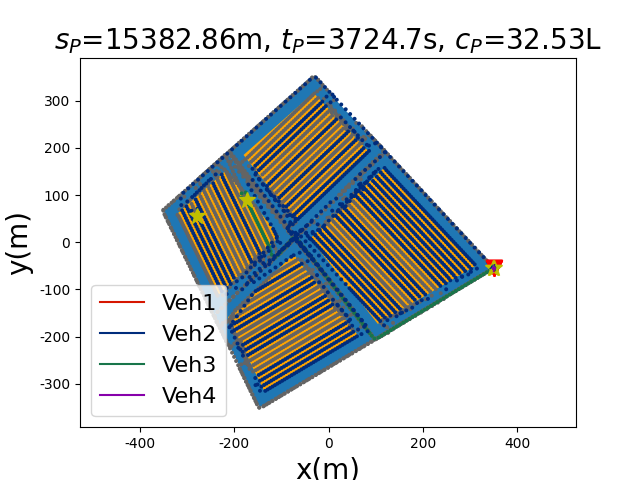}
        \caption{\small OGA-s}
    \end{subfigure}
    \hfill 
    \begin{subfigure}{0.325\textwidth}
        \centering
        \includegraphics[width=\linewidth]{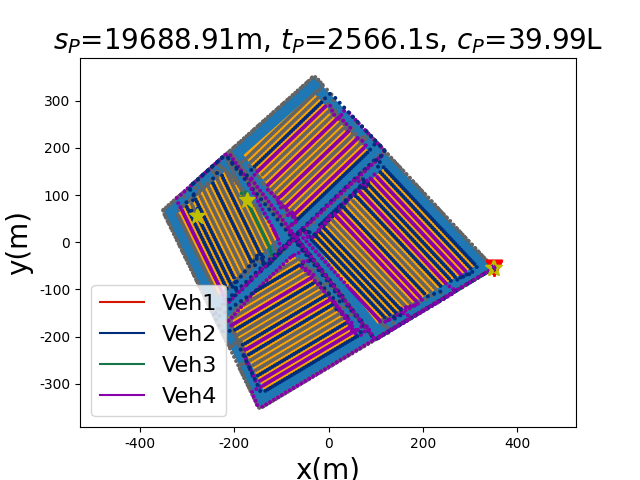}
        \caption{\small OGA-t}
    \end{subfigure}
    \hfill 
    \begin{subfigure}{0.325\textwidth}
        \centering
        \includegraphics[width=\linewidth]{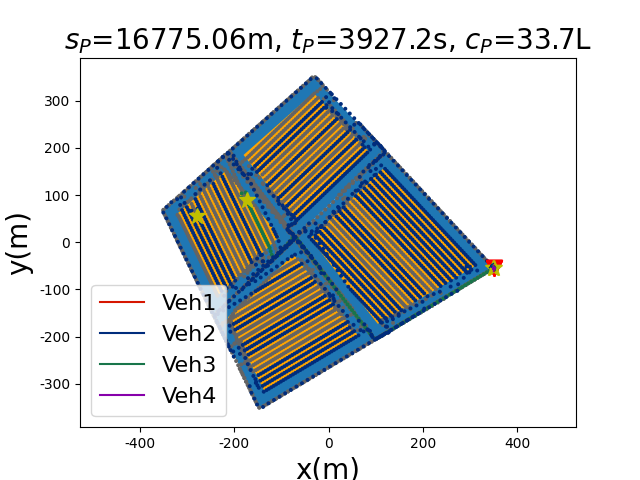}
        \caption{\small OGA-c}
    \end{subfigure}
    \caption{Simulation results of vehicle decrease task. In the first phase, all vehicles operate across the entire field. At 25\% of the operation time, Veh2 and Veh3 are within working lines, while Veh1 and Veh4 have returned to the depot. In the second phase, Veh1 remains at the depot, Veh3 returns directly to the depot, and Veh2 and Veh4 complete the remaining operational tasks.}
    \label{fig:case_study_veh}
\end{figure*}

Fig.\ref{fig:case_study_field} and Fig.\ref{fig:case_study_veh} illustrate JPDS-NN's performance across two tasks for three optimization objectives, using a test field with 4 plots and 4 vehicles. The first phase paths are shown as gray trajectories, second phase starts with yellow stars, and the depot is indicated by a red triangle.

It can be observed from Fig.\ref{fig:case_study_field} and Fig.\ref{fig:case_study_veh} that JPDS-NN demonstrates superior optimization in terms of distance efficiency, with vehicles exhibiting less detouring and fewer repetitive paths. Each vehicle essentially enters and exits the same plot only once, resulting in shorter overall travel distances across all samples. Comparing the optimization results of OGA and JPDS-NN with respect to time and fuel consumption, the trained neural network is capable of allocating tasks based on the vehicles' speed and fuel efficiency. In tests optimized for time, the network assigned more tasks to the faster vehicles; in tests optimized for fuel consumption, it allocated tasks to Veh2 and Veh3, which have lower operational fuel consumption. Furthermore, in the first and third columns of Fig.\ref{fig:case_study_veh}, although both JPDS-NN and OGA selected only Veh2 for the tasks, JPDS-NN's performance was better, indicating that JPDS-NN can provide a more optimized sequence of task allocations.

\section{CONCLUSIONS}

In conclusion, this study presents a novel approach to the Entrance Dependent Vehicle Routing Problem (EDVRP) in agricultural settings, addressing the complexities of routing multi-parameter vehicles within irregular fields. By introducing the Joint Probability Distribution Sampling Neural Network (JPDS-NN), we demonstrate a significant advancement in the efficiency and effectiveness of task allocation and route optimization. Our method not only leverages reinforcement learning for rapid end-to-end planning but also incorporates a pre-training task and a tailored reward combination strategy that enhance training performance and solution quality.

The experimental results affirm that JPDS-NN outperforms existing baseline methods, particularly in optimizing operational distance and fuel consumption. These findings underscore the potential of neural network-based solutions in transforming agricultural machinery routing, offering scalable and adaptive strategies that can meet the dynamic demands of farm operations. Future work will focus on further refining these algorithms and exploring their applicability in other complex routing scenarios, thereby contributing to the broader field of intelligent transportation and logistics in agriculture.

\section*{Acknowledgement}

This research was funded by National Science and Technology Major Project (2021ZD0110900).

\addtolength{\textheight}{-12cm}   


\bibliographystyle{ieeetr}
\bibliography{ref}

\end{document}